\documentclass[conference]{ieeeconf}  

\IEEEoverridecommandlockouts                              
\usepackage{graphicx} 
\usepackage{bm}
\usepackage{hyperref}

\title{\LARGE \bf
The Effect of Flagella Stiffness on the Locomotion of a Multi-Flagellated Robot at Low Reynolds Environment
}
 
\author{Nnamdi	Chikere, Yasemin Ozkan-Aydin,\textit{ Member, IEEE}
\thanks{$^{}$This work was not supported by any organization. All the authors are with the Department of Electrical Engineering, University of Notre Dame, Notre Dame, IN 46556 USA
        {\tt\small nchikere,yozkanay@nd.edu}}
}

\begin{document}
\maketitle
\thispagestyle{empty}
\pagestyle{empty}
\begin{abstract}
Microorganisms such as algae and bacteria move in a viscous environment with extremely low Reynolds ($Re$), where the viscous drag dominates the inertial forces. They have adapted to this environment by developing specialized features such as whole-body deformations and flexible structures such as flagella (with various shapes, sizes, and numbers) that break the symmetry during the motion. In this study, we hypothesize that the changes in the flexibility of the flagella during a cycle of movement impact locomotion dynamics of flagellated locomotion. To test our hypothesis, we developed an autonomous, self-propelled robot with four flexible, multi-segmented flagella actuated together by a single DC motor. The stiffness of the flagella during the locomotion is controlled via a cable-driven mechanism attached to the center of the robot. Experimental assessments of the robot's swimming demonstrate that increasing the flexibility of the flagella during recovery stroke and reducing the flexibility during power stroke improves the swimming performance of the robot. Our results give insight into how these microorganisms manipulate their biological features to propel themselves in low viscous media and are of great interest to biomedical and research applications.

\end{abstract}
\section{Introduction}

Understanding how microorganisms move in viscous environments has been of great interest to researchers due to its potential applications in solving microrobot locomotion challenges and designing better robotic swimmers. Microrobots exhibit immense potential for use in various domains, including biomedical applications for precision and minimally invasive surgeries and therapies \cite{ullrich_mobility_2013, diller_micro-scale_2013, sitti_biomedical_2015, kozielski_nonresonant_2021, nelson_microrobots_2010}, telemetry and sensing \cite{ergeneman_magnetically_2008, aziz_medical_2020}, targeted drug deliveries \cite{wang_microrobots_2023, gao_synthetic_2014, dogangil_toward_2008, lee_multifunctional_nodate}, and biopsy\cite{yim_biopsy_2014}. Furthermore, these microscale robots are suitable for environmental exploration and monitoring \cite{dubowsky_concept_2005, yoshimitsu_micro-hopping_2003, chen_biologically_2017} and other underwater exploratory applications \cite{yoshida_marangoni-propulsion_2022}. While these robots have great prospects for use, their progress has been limited by several challenges \cite{sitti_microscale_2007, diller_micro-scale_2013, dragomir_microrobotics_2014} such as lack of efficient manufacturing methods \cite{qin_micro-forming_2006}, power generation and storage in small size \cite{wang_moisture_2021}, and capacity to adapt to changing physical environments \cite{huang_adaptive_2019}. Finding an effective locomotion strategy is also crucial for microswimmers, and scientists have turned to microorganisms for inspiration in building robotic swimmers \cite{purcell_efficiency_1997, zhang_artificial_2010, huang_adaptive_2019, begey_manipulability_2020, khalil_controllable_2018, }. 
\begin{figure}[t]
    \centering
  \includegraphics[width=0.5\textwidth]{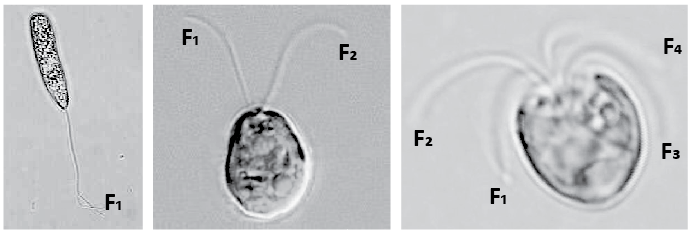}
    \caption{\textbf{Biological flagellated systems.} Flagellated algae with a different number of flagella (left) single, (middle) two, (right) four flagella. $F_i$ shows the i$^{th}$ flagella.}
    \label{fig:algaeFlagella}
\end{figure}

 \begin{figure*}[!h]
    \centering
    \includegraphics[width=0.8\textwidth]{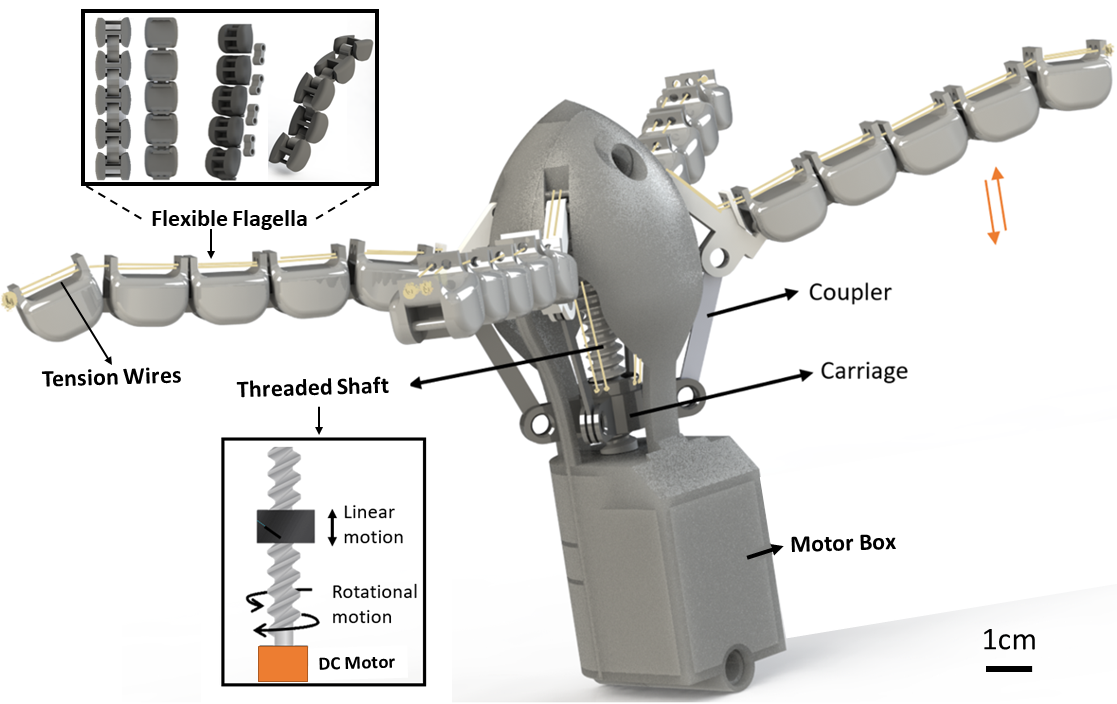}
     \caption{
     \textbf{Flagellated robot parts.} A computer-aided design drawing of a quadruped robot highlighting all main components. The flexible flagella subfigure shows that it is composed of short 3D-printed blocks connected by links. The threaded shaft subfigure shows the changes in types of motion during the actuation.
     }
     \label{fig:y_robotParts}
\end{figure*}

Cellular organisms such as bacteria, algae, and sperm cells thrive in a low Reynolds number environment - where viscous damping and fluid friction dominate inertia  \cite{purcell_life_1977,cohen_swimming_2010}. To achieve motion in such environments, we consider the "scallop theorem," which states that in a low Reynolds number fluid, the swimmer cannot achieve net displacement if the pattern of the motion is the same in both the forward and reverse directions (reciprocating motion) \cite{purcell_life_1977}. Therefore, as noted by Purcell \cite{purcell_life_1977}, to achieve net displacement, microswimmers need to change the geometrical sequence of the body (or appendages) shape and break the symmetry.

Microorganisms use physical adaptations like flagella or cilia to thrive in highly viscous habitats \cite{ghanbari_bioinspired_2020}. Flagella or cilia can be arranged in specific patterns on the microorganism’s body, allowing them to move in specific directions and break symmetry, which is a necessary condition to swim in such environments \cite{wan_synchrony_2020}. For instance, bacteria such as \textit{E. coli} and \textit{S. marcescens} \cite{nakamura_flagella-driven_2019} move by rotating helical flagella, while eukaryotic swimmers such as sperm cells \cite{wang_flagellar_nodate} and algae are propelled using a whip-like flagella structure \cite{elgeti_physics_2015}. Researchers are interested in understanding how microorganisms use these physical adaptations to thrive in their habitats because it provides insights into developing the most efficient robotic swimmers, particularly micro-robots, which would operate in low Reynolds environments like blood or bodily fluids \cite{rogowski_heterogeneously_2020, alapan_multifunctional_2020}.

Several previous studies \cite{dreyfus_microscopic_2005, singleton_micro-scale_2011, du_modeling_2022, ye_rotating_2014, du_simple_2021, diaz_minimal_2021} have highlighted desired features for the most efficient robotic swimmers which include being lightweight and small in size, able to move effectively and navigate within a specific environment, withstanding harsh environments or physical stresses, ability to be controlled and is equipped with relevant tools to gather information about their environment and communicate with other devices.

It can be noted that two requirements must be met for artificial micro-structures to move or swim under control \cite{cohen_swimming_2010, purcell_life_1977}. The device should first be mechanically deformable by the injection and transmission of energy (flexibility), and the order of these deformations also has to be cyclic and not time-reversible. The number of flagella and swimming gaits are particular features of interest that have been noted to affect their swimming performance \cite{hof_magnetic_nodate, diaz_minimal_2021}. 

In this study, we present an autonomous self-propelled robot with four multi-segmented flagellates actuated together by a single DC motor. We are particularly interested in how the changes in the flexibility of the flagella during stroke facilitate locomotion in low-Reynold numbers. There have been prior experimental studies into the flexibility of robotic swimmers, with researchers investigating the locomotion of robots with fully flexible flagella, passively flexible flagella \cite{diaz_minimal_2021}, and fully rigid flagella \cite{du_simple_2021, diaz_minimal_2021}. Our study shows a new approach inspired by the flagella beating wavelike pattern of microorganisms where we control the flexibility of the flagella during each cycle of motion and compare its effects on the swimming performance of the robot and the performance achieved with fully flexible flagella configurations.


 \begin{figure}[!t]
    \centering
     \includegraphics[width=0.5\textwidth,height=0.9\textheight,keepaspectratio]{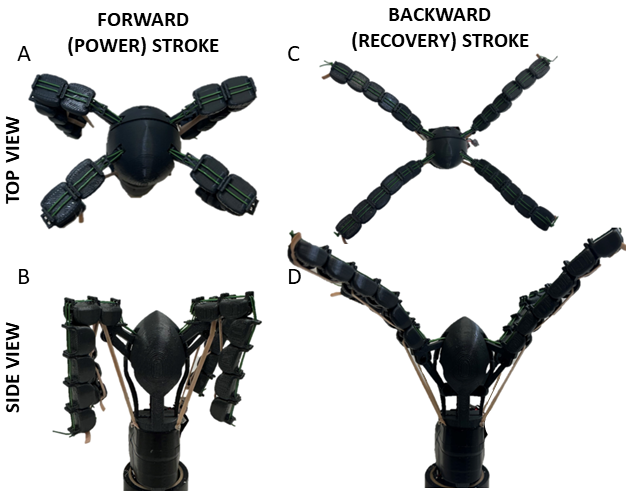}

     \caption{
     \textbf{States of controlled-flexible flagella}. A fully flexible state A (top view), and B (side view). The carriage is driven upward, causing the wires to be loose around the flagella links.
     B. A fully extended and rigid state, C (top view), and D (side view). The downward motion of the carriage pulls the wires tight, causing the flagellates to be extended and rigid.
     }  
     \label{fig:y_flagellaStates}
\end{figure}
\section{Materials and Methods}
 \subsection{Biological Inspiration}
    The quadriflagellate algae shown in Fig.\ref{fig:algaeFlagella}-(right) served as the biological inspiration for the robot. The organism has four flagella that extend from one end of its cell body and are aligned in two antiparallel pairs \cite{wan_synchrony_2020}. The algae swim by whiplike propulsions in a power and recovery stroke. taking on various gait patterns such as the pronk, gallop, and tronk gaits\cite{wan_synchrony_2020}. It can change the shape of its flagella during the strokes of motion, allowing it to generate complex swimming patterns and adapt to different environmental conditions. 
    
    Replicating the pronk gait swimming behavior of the quadriflagellate algae, we designed and built a mesoscale, underactuated, low-cost, 3D-printed robot that can traverse through highly viscous fluids by utilizing the controlled flexibility of its flagella.
    
  
     \begin{figure}[!t]
    \centering  \includegraphics[width=0.4\textwidth]{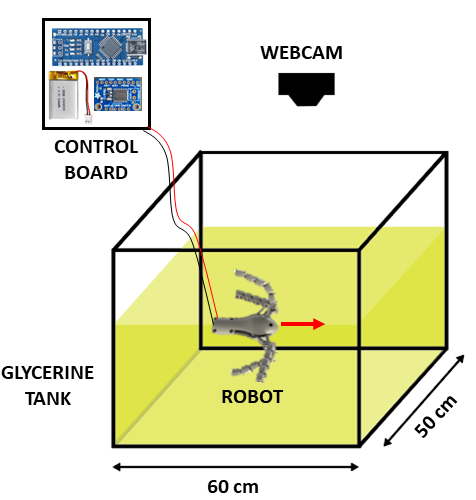}
     \caption{
     \textbf{Experimental setup} The setup consists of a transparent square tank filled with glycerine of high viscosity to approximate the low Reynolds number fluid regime experienced by the algae. The tank is fitted with an overhead digital camera to record the robot's dynamics. The robot is controlled via  custom control boars that include Arduino Nano and a motor driver.
     \label{fig:Experimentalsetup}
     }    
    \end{figure}  
 \subsection{Mechanical Design and Fabrication}
     The robot consists of a body (length = 12.6cm, weight = 125g) with a compartment housing the DC motor, a carriage and thread, and the flagellates (Fig. \ref{fig:y_robotParts}).  It was designed by Solidworks software, and 3D printed using Acrylonitrile Butadiene Styrene (ABS) with a Stratasys F170 printer. 
         \begin{figure*}[!h]
        \centering
         \includegraphics[width=1\textwidth,height=0.9\textheight,keepaspectratio]{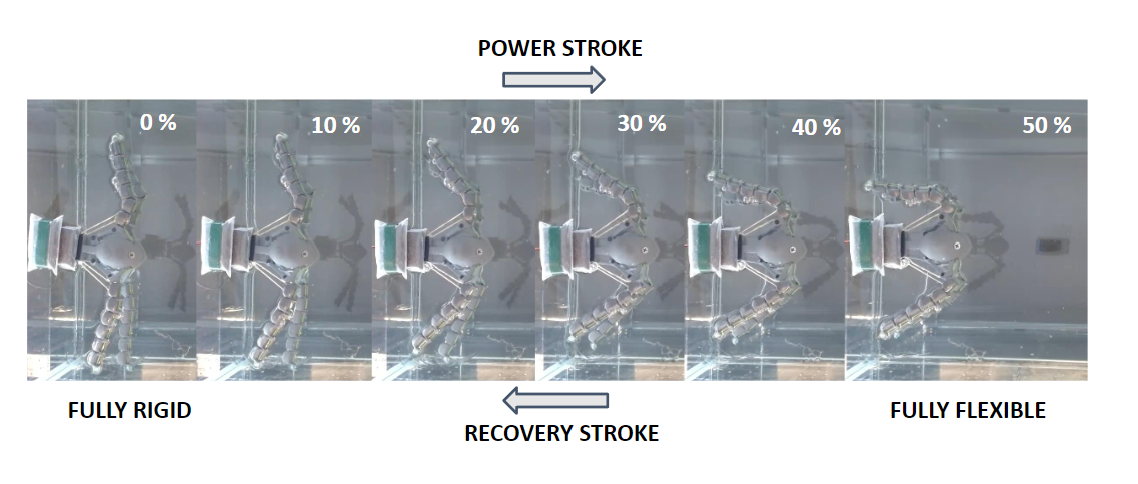}
         \caption{  
         \textbf{States of flagella.}The deformation of the controlled flagella during the cycle of motion (power and recovery stroke) in glycerine; the flagella are fully flexible at the end of a power stroke (50\% of a cycle) and fully rigid at the end of a recovery stroke (0\%  or 100\% of a cycle). 
         A small sheet of styrofoam (6cm by 4cm) was attached to the base section of the robot to improve its buoyancy.
         \label{fig:deform}
         }    
    \end{figure*} 
     The design of the robot was optimized to minimize the impact of hydrodynamic drag by implementing a pointed tip and a streamlined body (Fig. \ref{fig:y_robotParts}). Additionally, we integrated styrofoam strips along the robot's frame to enhance its buoyancy (Fig. \ref{fig:deform}). To enable unified control of the flagella with a single motor, we devised a simple mechanism that transformed the rotational motion of the motor into the linear motion of the carriage. This mechanism, depicted in the inset figure of Fig.\ref{fig:y_robotParts}, utilized a 3D printed boltlike carriage and a threaded shaft (Metric die M25 X 2.5, diameter = 8mm, length = 53mm, angle = $0^\circ$) linked to a DC motor (6 Volts 100 RPM, 70 mA Polulu DC gear motor with high torque). The selection of this particular motor was based on its ability to provide adequate torque for driving the shaft and cables while maintaining a sufficient propulsion speed in glycerine.

   The flagella comprise six 3D-printed components attached by short links in hinge joints. Incorporating hinge-type jointed parts and short links between the six 3D-printed components of the flagella introduces small gaps, resulting in passive flexibility. However, when the carriage pulls the wires down (Fig.\ref{fig:y_robotParts}), the gaps between the jointed parts close, causing the flagella to become rigid. This mechanism enables the robot to change the shape and flexibility of its flagella during motion cycles. A schematic of this mechanism is shown in Figure \ref{fig:y_flagellaStates}.
    In Fig.\ref{fig:y_flagellaStates}A-B, the flagella are fully flexible; the carriage has been driven to the upward point of the threaded shaft, and hence more wires are released into the jointed parts causing it to be flexible. In Fig. \ref{fig:y_flagellaStates}C-D, the carriage is driven downward, causing the wires to be pulled tight and the flagella rigid. These repetitive changes in the stiffness of the flagella alter the way the flagella interact with their environment by breaking symmetry, allowing them to generate net propulsive force  that propels the system forward. In order to augment the symmetry-breaking mechanism during the power stroke (Fig.\ref{fig:y_robotParts}), we utilized low-stiffness rubber bands (Office Depot® Brand Rubber Bands, '\#33') to connect the four top edges of the motor box to the connecting link located in the midsection of the flagella. During the rigid recovery stroke, the rubber bands are stretched out, whereas, during the power stroke, they contract to their original shape, inducing a flexible but bent shape that results in the desired symmetry-breaking effect.

     
     \begin{figure}[!h]

        \centering
       \includegraphics[width=0.5\textwidth, height=0.9\textheight,keepaspectratio]{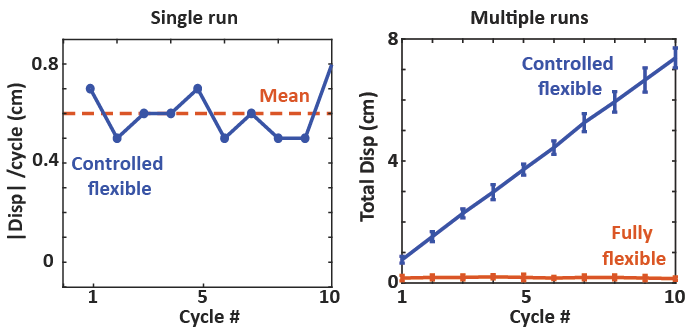}    

         \caption{  
         \textbf{Gait kinematics and performance of the robot.} Displacement of the robot in each motion cycle over 10 cycles (left). The left plot shows the average displacement that occurs in a cycle. Data was recorded for 10 cycles of motion.
         Total displacement for five runs of experiment over 10 cycles (right). This plot shows the total displacement of the robot over 10 cycles. The data was recorded for five runs of the experiment
          \label{fig:disp}}  
    \end{figure} 
 \subsection{Electronics}
     The four flagella are driven by a single DC gear motor (Polulu, 6 Volts, 100 RPM, 70 mA) coupled to a threaded rod with a carriage attached. This mechanism allows the rotational motion of the motor to be converted to a linear movement of the carriage. On the carriage, 28 AWG insulated copper wires are linked and connected from the carriage in the robot body to the flagella. The DC motor is driven by an Adafruit TB6612 1.2A DC/Stepper Motor Driver, which is then connected to an Arduino Pro Mini controller. The DC motor operates at 100RPM, and to achieve the desired clockwise and anticlockwise rotation of the motor's shaft, the Arduino was programmed to change the direction of motion of the robot in a five seconds sequence.
     During experimental trials, the Arduino controller was powered by a 5V USB power source, while a 6V DC power supply powered the DC motor.  Forward and reverse rotation of the motor induces upward and downward motion of the carriage, causing the cables attached to the flagellates to be taut and flagella rigid during the downward motion(recovery stroke) and flagellates flexible, and cables loose during the upward motion(power stroke). 

  \subsection{Experimental Setup}

  To ensure dynamic similarity with organisms operating at a microscale, we used vegetable glycerine (Bulk Apothecary\copyright) with a viscosity of 1.49 $Pa.s$, equivalent to a viscosity 1500 times greater than that of water and a density of 1000 $kg/m_3$ at 20°C. As shown in Fig.\ref{fig:Experimentalsetup}, a transparent tank (60x50x45 cm$^3$, CLEAR 2420, Waterbox Aquariums\copyright)  was filled with vegetable glycerine up to two-thirds of its volume. To monitor the robot's dynamics, an overhead camera (Logitech C920x HD Pro) was attached to the frame on the top of the tank. To conduct experiments, the robot is powered and placed near one end of the tank. The camera is then activated, capturing video footage of the robot's motion through the fluid at 30 fps.
  

\section{Results}

    In experiments, we tested two different types of robot flagella: fully-flexible and controlled-flexible flagella. 
    The fully flexible flagella configuration is the same as the controlled-flexible flagella given in Fig. \ref{fig:y_robotParts} (made from 3D printed links), but the parts are not connected with tension wires; hence the flexibility is not controlled. 
    We carried out experiments where we compared the performance of the different flagella configurations as a function of the displacement per cycle. The experiments for each configuration were carried out in five runs for 10 cycles each resulting in a total of 50 cycles. This was done to ensure the repeatability of the results, and a webcam recorded each run. 
   
    Multiple experimental tests were conducted on fully flexible flagella, and the results indicated negligible net displacement due to the flagella's inability to generate adequate thrust force to counteract the drag force and propel the robot (supplementary video). The fully flexible configuration exhibited results similar to those of the fully rigid, unhinged flagella configuration in \cite{diaz_minimal_2021}, where the robot could produce sufficient thrust force to counteract drag and achieve forward propulsion in the half cycle but returned to its initial position at the end of the cycle. This resulted from its similar motion pattern in both directions, which failed to break the time symmetry, leading to no net displacement over several cycles.
    
    The controlled-flexible configuration effectively generated sufficient thrust force during the power and recovery strokes, reducing the drag and enabling the robot to push through the viscous fluid (supplementary video). Additionally, the asymmetrical shape changes during the two strokes break the time symmetry of motion, resulting in a net displacement of the robot. 
\begin{figure}[!h]
\centering
\includegraphics[width=0.5\textwidth, height=0.9\textheight,keepaspectratio]
        {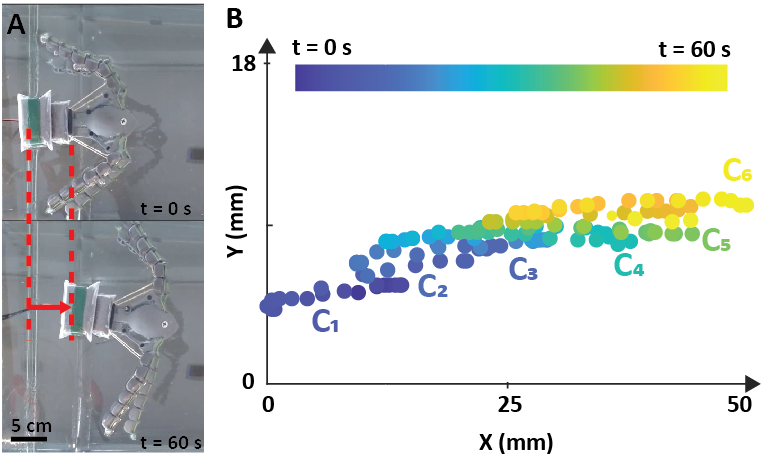}
    \caption{ \textbf{Tip trajectory over six cycles}. The plot shows the displacement of the robot's tip  during the six motion cycles on the direction of red arrow. The robot traveled a distance of 5cm in six cycles (C$_{1:6}$) taking a total time of 60 seconds (colorbar represents time).
          \label{fig:disp2}}  
    \end{figure}  
    In Fig. \ref{fig:disp}, we plot the displacement per cycle and the total displacement from experiments with the robot fitted with the controlled flexible and fully flexible flagella. The displacement per cycle plot in Fig. \ref{fig:disp}A shows the displacement of the robot over 10 cycles in a single run when in the controlled flexible configuration. We observed an average displacement rate of 0.7 ${\displaystyle \pm }$0.11 cm/cycle for each motion cycle. This means that in a complete cycle of motion, the robot displaced in the fluid by about 0.7 cm. In Fig. \ref{fig:disp}B, the plot shows the total displacement of the robot over 10 cycles using the controlled flexible flagella and fully flexible flagella for five different runs. The plot shows that in the controlled flexible configuration, the robot displaces by 7${\displaystyle \pm }$0.74 cm over 10 complete cycles. While in the fully flexible, the displacement is negligible.

    In Fig. \ref{fig:disp2}, we can observe the trajectory of the robot's tip over six cycles; the robot covered a distance of about 5 cm over 60 seconds. 
    The plot of displacement per cycle for five runs of the experiment with the controlled flexible configuration shows a linear relationship between the number of cycles and the displacement. This shows that our robot fulfilled the requirements for locomotion in low Reynolds number fluids as it breaks the time symmetry through shape change and produces enough thrust to propel the robot through the fluid.
    
    

\section{Conclusion}
    This paper proposes a novel method for improving the locomotion dynamics of low Reynolds number artificial swimmers via stiffness control. We introduced a quadriflagellate algae-inspired robot that changes the stiffness of its flagella during each motion cycle to break time symmetry and generate a non-reciprocal motion, a necessary condition for locomotion in low Reynolds number fluids. To control the stiffness of the flagella, we used a cable-driven mechanism coupled with a DC motor that changes the flexibility of the multisegmented flagella during each motion cycle. Experiments show that the changes in flexibility allow the robot to achieve a more efficient swimming performance and propel itself through a viscous fluid at a rate of 0.7${\displaystyle \pm }$0.11 cm/cycle compared to our robot configuration with the fully flexible flagella. 

    This research sheds some light on the intricate mechanics of microorganism locomotion, particularly in highly viscous fluids, and has provided valuable insights into the role of flagella stiffness in optimizing swimming performance. By mimicking the swimming behavior of microorganisms, we can develop robots that can autonomously navigate complex environments, perform specific tasks, and potentially aid in biomedical research applications and environmental explorations. The study of microorganism locomotion is a critical area of research in biology, and our findings have broad implications for understanding and advancing the field. The use of robotics and bio-inspired design has enormous potential for solving real-world challenges, and this research is an essential step in realizing that potential. 
    In the future, we plan to implement a more comprehensive control system for the robot's flagella stiffness, using different actuators to experiment with other gait patterns of microorganisms. By varying the stiffness of each flagellum independently, we can investigate how different combinations of flagellar motions affect the robot's overall swimming behavior. Additionally, by exploring new gait patterns, we may discover novel ways microorganisms navigate complex fluid environments, providing further insights into the mechanics of biological locomotion. Ultimately, this research will contribute to developing more advanced and efficient swimming robots with a wide range of potential applications.

\section{Acknowledgments}
 We thank the members of the MiNiRo-Lab at the University of Notre Dame for their help, support, and advice.

\bibliographystyle{ieeetr}

\end{document}